\documentclass[sigconf,natbib]{acmart}
\usepackage{multirow}
\usepackage{tcolorbox}
\usepackage{tabularx}
\usepackage{makecell}
\usepackage{siunitx}
\usepackage{booktabs}
\usepackage{float}
\usepackage{graphicx}
\usepackage{caption}
\usepackage{subcaption} % for subfigures
\usepackage{adjustbox}
\usepackage{enumitem}
\usepackage{arydshln}
\usepackage{balance}
\AtBeginDocument{%
  \providecommand\BibTeX{{%
    \normalfont B\kern-0.5em{\scshape i\kern-0.25em b}\kern-0.8em\TeX}}}

\copyrightyear{2025}
\acmYear{2025}
\setcopyright{cc}
\setcctype{by-nc}
\acmConference[SIGIR '25]{Proceedings of the 48th International ACM SIGIR Conference on Research and Development in Information Retrieval}{July 13--18, 2025}{Padua, Italy}
\acmBooktitle{Proceedings of the 48th International ACM SIGIR Conference on Research and Development in Information Retrieval (SIGIR '25), July 13--18, 2025, Padua, Italy}\acmDOI{10.1145/3726302.3729931}
\acmISBN{979-8-4007-1592-1/2025/07}

\begin{document}

%%
%% The ``title`` command has an optional parameter,
%% allowing the author to define a ``short title`` to be used in page headers.
\title{Combining Evidence and Reasoning for Biomedical Fact-Checking}

%%
%% The ``author`` command and its associated commands are used to define
%% ``authornote`` and ``authornotemark`` commands
\author{Mariano Barone}
\email{mariano.barone@studenti.unina.it}
\orcid{0009-0004-0744-2386}
\affiliation{%
  \institution{University of Naples Federico II}
  \city{Naples}
  \state{Italy}
  \country{Italy}
}

\author{Antonio Romano}
\email{antonio.romano5@unina.it}
\orcid{0009-0000-5377-5051}
\affiliation{%
  \institution{University of Naples Federico II}
  \city{Naples}
  \state{Italy}
  \country{Italy}
}

\author{Giuseppe Riccio}
\email{giuseppe.riccio3@unina.it}
\orcid{0009-0002-8613-1126}
\affiliation{%
  \institution{University of Naples Federico II}
  \city{Naples}
  \state{Italy}
  \country{Italy}
}

\author{Marco Postiglione}
\email{marco.postiglione@northwestern.edu}
\orcid{0000-0003-1470-8053}
\affiliation{%
  \institution{Northwestern University}
  \city{IL}
  \state{United States}
  \country{United States}
}

\author{Vincenzo Moscato}
\email{vincenzo.moscato@unina.it}
\orcid{0000-0002-0754-7696}
\affiliation{%
  \institution{University of Naples Federico II}
  \city{Naples}
  \state{Italy}
  \country{Italy}
}

\renewcommand{\shortauthors}{Mariano Barone, Antonio Romano, Giuseppe Riccio, Marco Postiglione, \& Vincenzo Moscato}

\begin{abstract}
Misinformation in healthcare, from vaccine hesitancy to unproven treatments, poses  risks to public health and trust in medical systems. While machine learning and natural language processing have advanced automated fact-checking, validating biomedical claims remains uniquely challenging due to complex terminology, the need for domain expertise, and the critical importance of grounding in scientific evidence. We introduce \textsf{CER} (Combining Evidence and Reasoning), a novel framework for biomedical fact-checking that integrates scientific evidence retrieval, reasoning via large language models, and supervised veracity prediction. By integrating the text-generation capabilities of large language models with advanced retrieval techniques for high-quality biomedical scientific evidence, \textsf{CER} effectively mitigates the risk of hallucinations, ensuring that generated outputs are grounded in verifiable, evidence-based sources. Evaluations on expert-annotated datasets (HealthFC, BioASQ-7b, SciFact) demonstrate state-of-the-art performance and promising cross-dataset generalization. Code and data are released for transparency and reproducibility: \url{https://github.com/PRAISELab-PicusLab/CER}.
\end{abstract}




\maketitle

\section{Introduction} \label{sec: introduction}

Distinguishing reliable information from misinformation represents a \emph{global challenge} that threatens the foundations of informed decision-making. Misinformation not only shapes public opinion and policy outcomes but also systematically erodes trust in authoritative information sources, thereby creating a recursive cycle that further complicates the pursuit of accurate knowledge \cite{doi:10.1177/2055207620948996}. The healthcare domain exemplifies the devastating impact of misinformation on public health outcomes. Misinformation surrounding evidence-based medical interventions --- including cancer treatments \cite{johnson2022cancer}, birth control\footnote{\url{https://www.washingtonpost.com/health/2024/03/21/stopping-birth-control-misinformation/}}, and other forms of medical care --- has led to devastating public health consequences, including treatment delays, increased mortality rates, and erosion of trust in healthcare systems \cite{gisondi2022deadly}. For example, during the COVID-19 pandemic, misinformation about vaccines led to  hesitancy, resulting in lower vaccination rates and increased spread of the virus. This misinformation was often propagated through social media platforms, where false claims about vaccine safety and efficacy spread rapidly \cite{DBLP:journals/tele/ApukeO21}. Additionally, misinformation about unproven treatments, such as the use of hydroxychloroquine, caused confusion and sometimes harmful self-medication practices \cite{saag2020misguided}.

\begin{figure}[t]
    \centering
    \includegraphics[width=.9\linewidth]{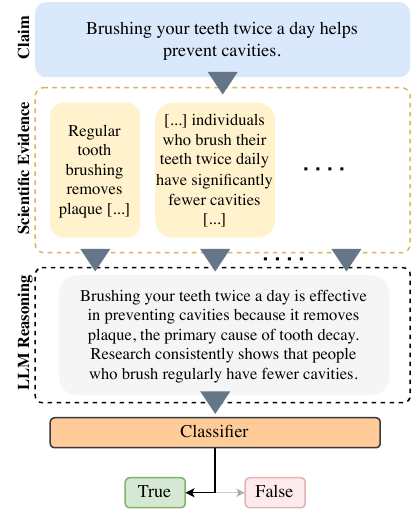}
    \caption{Combining Evidence and Reasoning (\textsf{CER}) for biomedical fact-checking. \textbf{(1)} Evidence supporting or opposing the biomedical claim is retrieved from scientific literature. \textbf{(2)} A large language model (LLM) is leveraged to justify the claim in light of the scientific evidence. \textbf{(3)} A classifier is trained to assess the veracity of the claim by leveraging the information collected in steps (1) and (2).}
    \label{fig:CER_wf_example}
    \Description{CER running example}
\end{figure}

Machine learning (ML) and natural language processing (NLP) techniques have emerged as powerful tools for automated fact-checking, demonstrating promising results across various domains \cite{DBLP:journals/ejis/SchuetzSV21,Operating_Machine}. While these computational approaches have been extensively studied and deployed in political fact-checking \cite{DBLP:conf/emnlp/RashkinCJVC17,DBLP:conf/sigir/VoL19,osti_10074110}, their effectiveness in validating biomedical claims remains limited. Biomedical fact-checking presents distinct challenges that transcend the capabilities of current automated systems. Unlike many other domains, medical claims often incorporate domain-specific terminology, rely on complex causal relationships, and require sophisticated interpretation of empirical evidence \cite{DBLP:journals/ijinfoman/Zhong23}. \emph{Furthermore, health-related misinformation demands particularly rigorous verification due to its potential impact on public health outcomes, necessitating both specialized domain knowledge and advanced reasoning capabilities that can contextualize claims within the broader framework of scientific evidence.}

Large language models (LLMs) have recently emerged as a promising approach to address these challenges in biomedical fact-checking. Their ability to process and reason about complex textual information, combined with their broad knowledge acquisition during pre-training, makes them potentially valuable tools for analyzing biomedical claims \cite{lui2024knowledge}. LLMs have demonstrated capabilities in understanding technical medical terminology, identifying relevant scientific evidence, and providing nuanced interpretations of research findings \cite{singhal2023large}. Retrieval-augmented generation (RAG) frameworks further enhance these capabilities by grounding LLM outputs in verified scientific literature, potentially mitigating the risk of hallucinations through direct access to peer-reviewed evidence \cite{kim2024vaiv,jeong2024improving}. \textit{By dynamically retrieving and incorporating relevant medical research during the analysis process, RAG systems could provide more reliable and verifiable fact-checking results}. However, these approaches still face important limitations that must be carefully considered: (1) LLMs, even when augmented with retrieval, can still generate plausible-sounding but factually incorrect information. (2) The quality of fact-checking remains heavily dependent on the relevance and reliability of the retrieved evidence, and challenges persist in efficiently identifying and incorporating the most pertinent scientific literature. (3) Additionally, their knowledge cutoff dates and potential biases in training data can lead to outdated or incomplete medical information. 

In this work, we propose a novel approach to biomedical fact-checking that integrates three key components: systematic scientific evidence retrieval, LLM reasoning capabilities, and supervised veracity prediction. Our system, \textsf{CER} (Combining Evidence and Reasoning), addresses the challenges in biomedical claim verification through a multi-stage architecture (illustrated in Figure \ref{fig:CER_wf_example}) that synthesizes (1) the input claim, (2) relevant scientific evidence from peer-reviewed literature, and (3) LLM-generated justifications. This architecture leverages the advanced reasoning capabilities of LLMs while maintaining factual grounding through explicit evidence linkage, thereby mitigating the risk of hallucinations. The proposed architecture follows an open-domain fact-checking paradigm, which eliminates reliance on a predefined set of evidence \cite{DBLP:conf/cvpr/AbdelnabiHF22}. Rather than being restricted to a fixed corpus of golden evidence, this approach dynamically retrieves relevant information from an extensive external knowledge base, enabling a \textit{content-driven} evaluation of claims. We evaluate \textsf{CER} on three benchmarks annotated by domain experts: HealthFC \cite{DBLP:conf/coling/VladikaSM24}, BioASQ-7b \cite{DBLP:journals/corr/abs-2006-09174}, and SciFact \cite{wadden2020emnlp}. The results show that \textsf{CER} consistently outperforms state-of-the-art methods, achieving up to a +3.43\% improvement in F1 scores. To provide deeper insights, we analyze the performance impact of different retrieval strategies (sparse vs. dense) and examine the contributions of scientific evidence integration and LLM-based reasoning. Additionally, cross-dataset evaluations highlight the generalizability and robustness of our framework across diverse settings. While these results underscore its effectiveness in healthcare, the modular design of \textsf{CER} makes it adaptable to other domains requiring domain-specific evidence and advanced reasoning. To promote transparency and reproducibility, we release the code, evidence, and outputs generated during our experiments\footnote{\textbf{Code is shared anonymously; data will be available upon publication.}}.

\section{Combining Evidence and Reasoning (\textsf{CER})} \label{sec: methodology}
% AGGIUNGERE UNA BREVE DESCRIZIONE DELL'ARCHITETTURA DETERMINANDO LE FASI CHE PORTI AVANTI DURANTE IL FACT-CHECKING

\begin{figure*}[ht]
    \centering \includegraphics[width=0.95\textwidth]{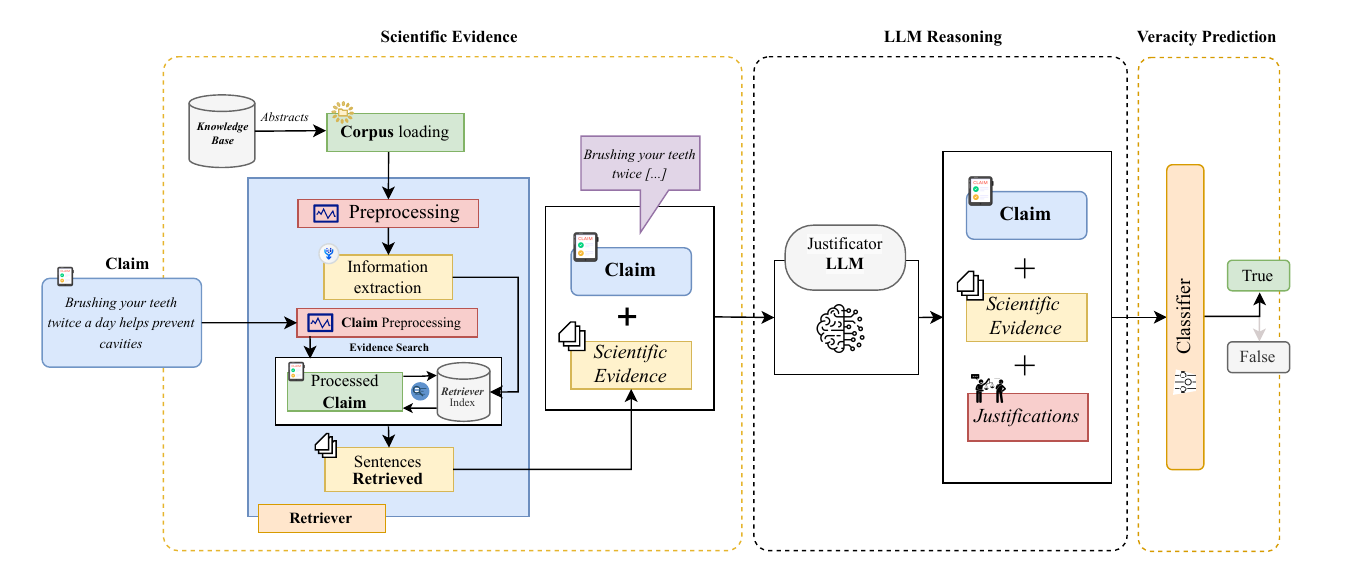}
    \caption{\textsf{CER} workflow. Claims are preprocessed before being matched against a scientific corpus indexed using a retrieval system. Relevant evidence passages are retrieved and concatenated with the original claim as input to a large language model (LLM), which generates detailed justifications. A binary classifier then evaluates the claim's veracity by analyzing the claim text, retrieved scientific evidence, and generated justifications to produce a final true/false assessment}
    \label{fig:CER_workflow}
    \Description{CER workflow}
\end{figure*}

In this section, we present our proposed methodology to classify biomedical \textit{claims}\footnote{In this work, a \textit{claim} refers to any statement or question whose truthfulness or factual accuracy can be evaluated based on external evidence. This broad definition encompasses declarative sentences as well as interrogative forms.}. 
Our architecture, as illustrated in Figure \ref{fig:CER_workflow}, is divided into three main components: (1) Scientific Evidence Retrieval, (2) LLM Reasoning and (3) Veracity Prediction. The remainder of this section provides an in-depth description of each component. 
% The first component focuses on \textbf{Scientific Evidence Retrieval} (1). This stage employs both dense and sparse retrievers to identify relevant evidence from extensive knowledge sources. The retrieved evidence is subsequently ranked based on relevance scores and filtered to retain only the most pertinent information. The performance of the retrieval module is evaluated using standard fact-checking datasets and metrics such as recall and mean average precision (MAP). The second focuses on \textbf{LLM Reasoning} (2). In this phase, the retrieved evidence is passed to a large language model (LLM) for reasoning and analysis. The LLM synthesizes the evidence, assesses its alignment with the claim, and generates comprehensive explanations. To balance generalization and domain specificity, we employ a combination of zero-shot and fine-tuned LLMs. The reasoning module uses prompts designed to encourage evidence-based analysis and are explicitly tailored for domain-specific claims. The third component involves \textbf{Veracity Prediction} (3). In this final stage, veracity is predicted using both zero-shot and fine-tuned LMs. These predictions are derived from the reasoning outcomes generated in the previous module. To ensure transparency and a thorough understanding of performance improvements, we conduct ablation studies to quantify the contribution of each module to the overall system performance.

% QUI VA LO SCHEMA - WORKFLOW del PROGETTO

\subsection{Scientific Evidence Retrieval} \label{sec: scientific_evidence}

The workflow starts with the Scientific Evidence Retrieval module, which interfaces with scientific knowledge bases to extract domain-specific claims. Similarly to previous studies \cite{DBLP:journals/corr/abs-2404-08359,Jin_2023,sarrouti-etal-2021-evidence-based}, we leverage PubMed\footnote{\url{https://zenodo.org/records/7849020}} as our primary knowledge source, given its comprehensive repository of peer-reviewed biomedical literature. We specifically focus on article abstracts, which offer concise yet comprehensive summaries of research findings, methodologies, and conclusions. This choice is motivated by both the information density of abstracts and the common restrictions on access to full-text articles, while providing sufficient context for validating medical claims.

\paragraph{Retrieval Module}. To extract sentences from the PubMed knowledge base, we employed Sparse Retrieval and Dense Retrieval strategies, both widely used in open-domain question-answering systems.
We utilized a BM25 inverted index \cite{robertson2009ftir} to rank documents based on term-matching relevance. The BM25 scoring function computes the relevance of a document $D$ to a query $Q$ as shown as follows: 
\[
\text{score}(Q, D) = \sum_{t \in Q} \text{IDF}(t) \cdot \frac{f(t, D) \cdot (k_1 + 1)}{f(t, D) + k_1 \cdot \left(1 - b + b \cdot \frac{|D|}{\text{avgdl}}\right)},
\]
where:
\begin{itemize}
    \item $\text{IDF}(t)$ is the inverse document frequency, calculated as:
    \[
    \text{IDF}(t) = \log \frac{N - n_t + 0.5}{n_t + 0.5},
    \]
    with $N$ as the total number of documents and $n_t$ as the number of documents containing the term $t$
    \item $f(t, D)$ is the term frequency of $t$ in document $D$
    \item $|D|$ is the length of the document $D$, and $\text{avgdl}$ is the average document length in the collection
    \item $k_1$ and $b$ are hyperparameters controlling term frequency saturation and document length normalization, respectively
\end{itemize}
Text preprocessing included normalization, tokenization, stemming, stopword removal, and punctuation standardization. The top $k$ results ($k = 20$) were ranked using cosine similarity over term-based features, offering efficiency and interpretability for term-centric queries. 
For Dense Retrieval, we employed the multi-qa-MiniLM-L6-cos-v1 \footnote{\url{https://huggingface.co/sentence-transformers/multi-qa-MiniLM-L6-cos-v1}}, a pre-trained SBERT model optimized for semantic similarity tasks. This model generates dense vector embeddings for textual inputs, capturing semantic nuances such as synonyms and paraphrases. The similarity between a query $Q$ and a document $D$ is measured using cosine similarity: $\text{sim}(Q, D) = \frac{\mathbf{q} \cdot \mathbf{d}}{\|\mathbf{q}\| \cdot \|\mathbf{d}\|}$, where $\mathbf{q}$ and  $\mathbf{d}$ are the embedding vectors of the query $Q$ and document $D$, respectively, and  $\|\mathbf{q}\|$ and $\|\mathbf{d}\|$ are their Euclidean norms.

\paragraph{Combining Claims and Scientific Evidence}. Once the knowledge base is indexed, claims from the dataset are processed and used as search queries. The retrieval system identifies relevant sentences from the indexed database by evaluating term-level relevance between the query and the stored text. This step ensures that the retrieved evidence is contextually aligned with the given claim. For each claim, the system extracts up to three pieces of evidence from the retrieved results. The extracted evidence is structured in the following format: ''\texttt{claim [SEP]} \texttt{\textit{evidence\_1, evidence\_2, evidence\_3.}}'' Here, \texttt{[SEP]} acts as a separator token to distinguish the claim from the evidence pieces, while the evidence sentences are concatenated with commas to maintain clarity and consistency in the input format. Based on preliminary experiments, using three pieces of evidence usually provides sufficient support without redundancy or excessive processing costs. The final claim-evidence pairs are stored in a structured format and prepared for the next stage of processing, where they will be used as input for reasoning by a large language model (LLM). This integration of retrieval and reasoning ensures a streamlined pipeline for claim verification tasks.
\subsection{LLM Reasoning}

\begin{comment}
\begin{table}[t]
    \centering
    \scalebox{0.95}{
    \begin{tabularx}{\columnwidth}{lX}
        \toprule
        \textbf{Dataset} & SciFact \\
        \midrule
        \textbf{Claim} & Antibiotic induced alterations in the gut microbiome increase resistance against Clostridium difficile. \\
        \midrule
        \textbf{Justification} & \textit{The provided context indicates that antibiotic-induced alterations in the gut microbiome do not increase resistance against Clostridium difficile. Instead, they reduce colonization resistance against pathogens, including Clostridium difficile. This disruption of the gut microbiota by antibiotics is a major risk factor for Clostridium difficile infection (CDI). The mechanisms by which antibiotics reduce colonization resistance against C. difficile are not fully understood, but they are important for the development of preventative and therapeutic approaches against this pathogen.} \\
        \bottomrule
    \end{tabularx}}
    \caption{Example of LLM reasoning output for a scientific claim, showing the generated justification based on retrieved evidence. The claim originates from the SciFact dataset \cite{wadden2020emnlp}, a benchmark for scientific claim verification.}
    \label{fig:LLM_Reasoning_Example}
\end{table}
\end{comment}

In this phase, \textsf{CER} leverages LLMs as reasoning assistants rather than autonomous decision-makers. This design choice explicitly mitigates the risk of hallucinations inherent in LLMs when used for standalone fact-checking. The LLM's role is twofold: (1) to assess the claim's \textit{veracity} based on the provided scientific evidence and (2) to generate a detailed \textit{justification} for this assessment. 
\begin{comment}
Figure \ref{fig:LLM_Reasoning_Example} shows an output example (additional examples are reported in Appendix).
\end{comment}
Both the binary prediction and the accompanying reasoning serve as structured inputs for the downstream Veracity Prediction module, which will determine the final classification. The LLM's analysis is guided by a prompt template that systematically combines the claim with the retrieved evidence (Table \ref{tab:SciFact} shows an example; additional examples are provided in Appendix \ref{sec:appendix-C}), as shown below: 
\begin{tcolorbox}[colback=blue!10, colframe=blue!30, title=LLM Resoning Prompt Structure]
    \textbf{Claim}: [claim]. 
    
    \textbf{Evidence}: [evidence 1, evidence 2, evidence 3]. 
    
    \textit{Provide a justification for your assessment about the veracity of the claim.}
\end{tcolorbox}

\begin{table}[t]
    \centering
    \scalebox{0.95}{
    \begin{tabularx}{\columnwidth}{lX}
        \toprule
        \textbf{Dataset} & SciFact \\
        \midrule
        \textbf{Claim} & A deficiency of folate increases blood levels of homocysteine. \\
        \midrule
        \textbf{Justification} & \textit{Folate deficiency has been shown to increase blood levels of homocysteine. This is due to the role of folate in the metabolism of homocysteine, where it acts as a coenzyme. In cases of folate deficiency, the remethylation pathway of homocysteine is inhibited, leading to its efflux into the circulation. This results in elevated homocysteine levels, which have been implicated in various adverse health conditions. Studies have reported a significant inverse correlation between plasma folate levels and circulating homocysteine levels, and a significant increase in plasma homocysteine concentrations in patients with marginal folate deficiency. Folate supplementation has been shown to reduce plasma homocysteine levels, even in patients with normal folate status and those with cobalamin deficiency.} \\
        \bottomrule
    \end{tabularx}}
    \caption{Example of LLM reasoning output for a scientific claim, showing the generated justification based on retrieved evidence. The claim originates from the SciFact dataset \cite{wadden2020emnlp}, a benchmark for scientific claim verification.}
    \label{tab:SciFact}
\end{table}

\subsection{Veracity Prediction}

While the LLM reasoning module generates structured analyses, relying solely on these outputs for final veracity decisions would be problematic due to potential reasoning gaps, inconsistencies between justifications and conclusions, or biases in the LLM's decision process. To address these limitations, we implement a dedicated verification layer that evaluates both the LLM's reasoning and the underlying evidence to produce more reliable classifications. 
Building upon established approaches in veracity prediction \cite{DBLP:journals/tkdd/VosoughiMR17}, we implement a classification framework that assigns one of three labels to each claim: ``true``, ``false``, or ``insufficient evidence``. We explore and compare two distinct methodological approaches --- i.e., zero-shot classification and fine-tuning --- for the veracity prediction task.

\paragraph{Zero-shot classification}. The first technique leverages a language model (LM) with zero-shot classification capabilities. In this approach, the LM processes claims alongside the extracted sentences and justifications, directly providing a classification output. Zero-shot classification refers to the ability of the model to generalize and perform tasks it has not been explicitly trained for. Formally, given a claim $C$ and evidence $E$, the model predicts a class label $y \in \{y_1, y_2, \dots, y_k\}$ using its pre-trained knowledge: $P(y \mid C, E) = \text{LM}(C, E)$, where $P(y \mid C, E)$ represents the probability distribution over the possible labels. 

This approach is particularly advantageous in scenarios where labeled data is unavailable, as it allows for rapid deployment without the need for additional task-specific training. The model utilizes its broad general knowledge acquired during pre-training to interpret the claim and evidence. However, the zero-shot approach has limitations in domain-specific tasks, where generalization may be inadequate for capturing nuanced or highly specialized information.

\paragraph{Fine-tuning}. The second technique involves fine-tuning the LM, adapting it to the specific task by retraining it on a smaller, domain-specific dataset. Fine-tuning is especially effective for specialized tasks where domain-specific knowledge is critical. During this process, the LM learns to predict a target label $y$ by minimizing the task-specific loss function $\mathcal{L}$ over the fine-tuning dataset $\mathcal{D}$: 
\[
\mathcal{L} = - \sum_{(C, E, y) \in \mathcal{D}} \log P(y \mid C, E),
\]
where $\mathcal{D}$ consists of tuples $(C, E, y)$, with $C$ as the claim, $E$ as the evidence, and $y$ as the correct label. 
In our case, we fine-tuned the model using claims, scientific evidence, and justifications generated in the preceding steps. This enabled the LM to learn directly from task-relevant examples, resulting in enhanced accuracy and improved fact-checking performance. Unlike the zero-shot approach, fine-tuning offers higher accuracy for domain-specific tasks but requires labeled data and additional computational resources, making it a more resource-intensive method.

\section{Experiments} \label{sec: experiments}

This section presents a comprehensive empirical evaluation of our proposed \textsf{CER} architecture across multiple datasets and experimental configurations. Our evaluation protocol encompasses four key objectives: (1) benchmarking \textsf{CER} against state-of-the-art closed- and open-source baselines, (2) conducting ablation studies to quantify the contribution of each pipeline component, (3) analyzing architectural variants through systematic comparison of retrieval strategies (dense vs. sparse), LLM reasoning approaches (diverse prompting techniques), and veracity prediction methods (zero-shot vs. fine-tuned classification), and (4) assessing the framework's generalization capabilities through cross-dataset evaluation.

\subsection{Experimental Setup}

\subsubsection{Datasets}
We evaluate \textsf{CER} on three datasets: HealthFC \cite{DBLP:conf/coling/VladikaSM24}, BioASQ-7b \cite{DBLP:journals/corr/abs-2006-09174}, and SciFact \cite{wadden2020emnlp}. {HealthFC} \cite{DBLP:conf/coling/VladikaSM24} is a dataset for question answering and fact-checking, comprising 750 health-related claims labeled as true, false, or NEI (not enough information), verified by medical professionals using scientific evidence. {BioASQ-7b} \cite{DBLP:journals/corr/abs-2006-09174} contains 745 biomedical claims from the BioASQ challenge, labeled as true or false, supporting the development of question-answering systems for biomedical queries. {SciFact} \cite{wadden2020emnlp} includes 1.4k expert-written scientific claims with evidence-based labels (``support``, ``confute``, or NEI) and rationales, focusing on validating claims through specialized domain knowledge. These datasets were annotated by domain experts, with rigorous validation of the assigned labels to ensure accuracy and reliability. This careful curation guarantees that the datasets reflect high-quality, validated medical knowledge. Table \ref{tab:datasets} presents details of label distributions.

\begin{table}[t]
    \centering
    \begin{tabular}{llccc}
        \hline
        \textbf{Dataset} & \textbf{Domain}         & \checkmark & \(\times\) & \textbf{?} \\ \hline
        \textbf{HealthFC} & Health-related online searches           & 202        & 125        & 433        \\
        \textbf{BioASQ-7b} & Biomedical research     & 614        & 131        & ---        \\
        \textbf{SciFact}   & Scientific claims          & 556        & 516        & 337        \\
        \hline
    \end{tabular}
    \caption{Overview of the datasets used in the experiments, including their domain and label distributions (\checkmark – true labels, \(\times\) – false labels, \textbf{?} – NEI labels).}
    \label{tab:datasets}
\end{table}

\subsubsection{Implementation and Training details}
To implement the Sparse and Dense Retriever, we utilized a system featuring an Intel(R) Core(TM) i7-9700K CPU @ 3.60GHz processor, 32GB of RAM, and a GT710 graphics card. All training experiments were conducted on a machine equipped with an Intel Xeon CPU with 2 vCPUs (virtual CPUs), 13GB of RAM, and an NVIDIA Tesla T4 GPU with 16GB of VRAM. The LLM used to generate justification in the ``LLM Reasoning`` module is \textit{Mixtral-8x22B-Instruct-v0.1} \cite{jiang2024mixtralexperts}. Its Mixture-of-Experts (MoE) design excels at managing complex language structures and synthesizing information across diverse contexts, making it particularly effective for domain-specific tasks like medical text synthesis. Comparative experiments with other LLMs, including those pre-trained on biomedical corpora, are detailed in the Appendix \ref{sec:appendix-B}. DeBERTa (``deberta-v3-large-zeroshot-v1.1-all-33``) \cite{he2023iclr}, BERT (``bert-base uncased``) \cite{DBLP:journals/corr/abs-1810-04805}, and PubMedBERT \cite{gu2022health} were tested for the implementation of the ``Veracity Prediction`` module.
\begin{figure}[t]
    \centering
    \begin{subfigure}[b]{0.32\linewidth}  % Imposta la larghezza a quasi la metà della linea
        \centering
        \includegraphics[width=\linewidth]{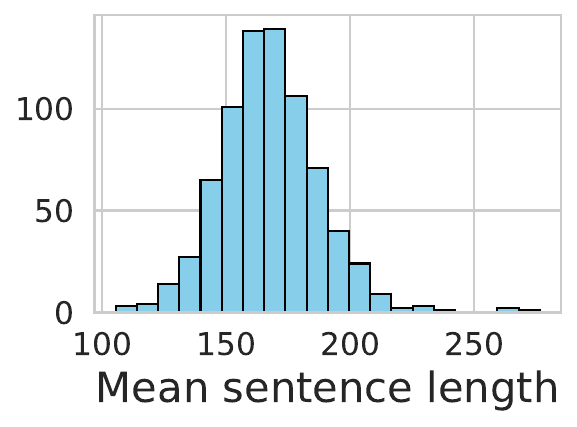}
        \caption{HealthFC}
        \label{fig:istogramma2}
    \end{subfigure}
    \hfill
    \begin{subfigure}[b]{0.32\linewidth}  % Imposta la larghezza a quasi la metà della linea
        \centering
        \includegraphics[width=\linewidth]{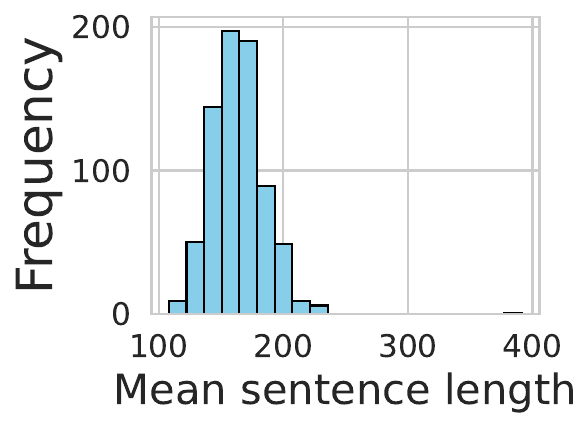}
        \caption{BioASQ-7b}
        \label{fig:istogramma1}
    \end{subfigure}    
    \label{fig:combined_istograms}
    \hfill
    \begin{subfigure}[b]{0.32\linewidth}  % Imposta la larghezza a quasi la metà della linea
        \centering
        \includegraphics[width=\linewidth]{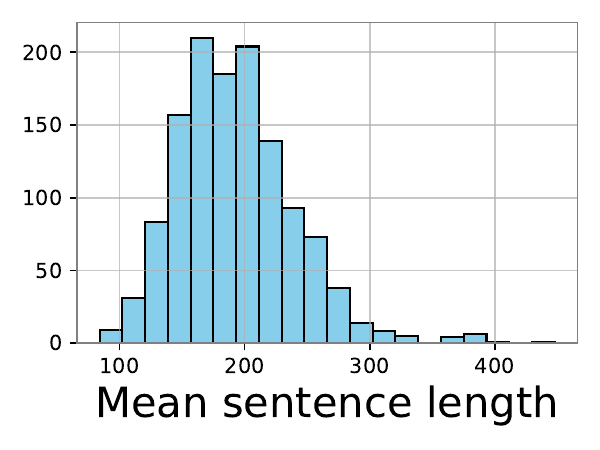}
        \caption{SciFact}
        \label{fig:istogramma3}
    \end{subfigure}    
    \caption{Average sentence length distributions of HealthFC (a), BioASQ-7b (b), SciFact (c). \textit{Mean sentence length} refers to the average number of characters per sentence in each documen}
    \label{fig:combined_istograms3}
    \Description{Avg sentence length distribution}
\end{figure}
Figure \ref{fig:combined_istograms3} illustrates the sentence length distributions for HealthFC, BioASQ-7b, and SciFact, emphasizing the complexity and variability of the datasets used. To address input token limitations, we use 20 sentences per claim in the ``Scientific Evidence`` step and 4 sentences in the ``Veracity Prediction`` step, maximizing contextual information for justification creation and label prediction. Veracity Prediction models have been trained using original training and validation datasets, with hyperparameters optimized through grid and random search to ensure generalization and prevent overfitting (details in Appendix \ref{sec:appendix-A}). Performance was assessed on the independent test set to ensure unbiased metrics, with training and test splits detailed in Table \ref{tab:fine-tuning division}. Evaluation relied on macro precision, macro recall, and macro F1 scores.

\begin{table}[t]
    \centering
    \scalebox{0.99}{ % Usa anche \resizebox{\columnwidth}{!} se preferisci
    \begin{tabular}{lccc}
        \hline
        \textbf{Dataset} & \textbf{Train} & \textbf{Validation} & \textbf{Test} \\
        \hline
        \textbf{HealthFC} & 451 & 151 & 151 \\
        \textbf{BioASQ-7b} & 447 & 150 & 150 \\
        \textbf{SciFact}  & 809 & 300 & 300 \\
        \hline
    \end{tabular}}
    \caption{Training, validation, and test set distributions of HealthFC, BioASQ-7b, and SciFact.}
    \label{tab:fine-tuning division}
\end{table}

\subsubsection{Baselines}
We compared \textsf{CER} with open-source and closed-source fact-checkers, as shown in Table \ref{tab:results_healthfc_bioasq}. Factiverse.AI and Perplexity.AI are closed-source platforms that categorize evidence as ``Supporting``, ``Mixed``, or ``Disputing`` using sources like Google, Bing, Wikipedia, and FactSearch, with flexibility in managing the ``Mixed`` label. In particular, for Perplexity.AI, each claim was submitted as a direct query, and the system's textual answer was manually mapped to three classes: responses expressing certainty and agreement were labeled as True, disagreement as False, and vague or inconclusive responses as Not Enough Information (NEI). \citet{DBLP:journals/corr/abs-2404-08359} proposes a voting-based veracity prediction architecture, leveraging sentences from knowledge bases, similar to our approach. \citet{10.1145/3485127} focuses on document and sentence retrieval from Wikipedia for claim validation. \citet{zaheer2020nips} employs BigBird, a sparse-attention transformer, while \citet{DBLP:journals/corr/abs-1909-11942} uses ALBERT for text classification, trained on datasets like FEVER, HOOVER, and FEVEROUS. \citet{DBLP:conf/coling/VladikaSM24} applies pipeline and joint systems for medical claim fact-checking. Simple baselines (``All True``, ``All False`` and ``All NEI``) were also included for comparison.

\begin{table*}[ht]
  \centering
  \begin{adjustbox}{width=\textwidth} % Scala la tabella per adattarla alla larghezza della pagina
    \begin{tabular}{llccccccccccc}
      \hline
      \textbf{Category} & \textbf{Fact-Checker} & \multicolumn{3}{c}{\textbf{HealthFC}} & & \multicolumn{3}{c}{\textbf{BioASQ-7b}} & & \multicolumn{3}{c}{\textbf{SciFact}} \\
      \cline{3-5} \cline{7-9} \cline{11-13}
                       &                         & \textbf{Precision (\%)} & \textbf{Recall (\%)} & \textbf{F1 (\%)} & & \textbf{Precision (\%)} & \textbf{Recall (\%)} & \textbf{F1 (\%)} & & \textbf{Precision (\%)} & \textbf{Recall (\%)} & \textbf{F1 (\%)} \\
      \hline
      Baselines         & All True (baseline)     & 8.97 & 33.33 & 14.14 & & 82.39 & 1.0 & 90.34 & & 13.71 & 33.33 & 19.43 \\
                       & All False (baseline)    & 5.55 & 33.33 & 9.52  & & 17.60 & 1.0 & 29.94 & & 7.16  & 33.33 & 11.78 \\
                       & All NEI (baseline)      & 18.79 & 33.33 & 24.04 & & -- & -- & -- & & 12.47 & 33.33 & 18.15 \\ \hline
      Online platforms        & Factiverse.AI \cite{DBLP:conf/sigir/Setty24a} & 53.47 & 30.19 & 38.59 & & 91.31 & 84.01 & 87.51 & & 50.25 & 46.80 & 48.46 \\
                       & Perplexity.AI\footnotemark[6] & 56.40 & 47.20 & 51.39 & & 83.20 & 80.13 & 81.64 & & 52.75 & 50.00 & 51.34 \\ \hline
      LLM Predictions  & Mixtral LLM prediction\footnotemark[7] & 40.10 & 41.24 & 40.61 & & 72.43 & 70.69  & 71.54 & & 45.89 & 40.60 & 43.02 \\
                       & GPT4o mini \cite{xie2024miniomni2opensourcegpt4ovision}             & 40.52 & 39.88 & 39.06  & &  53.85 & 52.71 & 52.79& & 39.01 & 37.64 & 38.31 \\
                        \hline
                    
      Literature baselines     
      & \citet{DBLP:journals/corr/abs-2404-08359} & 52.60 & 44.50 & 40.60 & & 60.90 & 64.40 & 61.70 & & 46.00 & 42.30	 & 44.07 \\
      & \citet{10.1145/3485127} & 41.72 & 49.36 & 45.21 & & 47.42 & 52.36 & 49.76 & & 54.12 & 51.20 & 52.62 \\
                       & \citet{zaheer2020nips} & 26.67 & 35.95 & 30.62 & & 69.62 & 33.57 & 45.29 & & 38.70 & 35.20 & 36.87 \\
                       & \citet{DBLP:journals/corr/abs-1909-11942} & 21.40 & 39.36 & 27.72 & & 74.60 & 50.37 & 60.13 & & 34.45 & 31.85 & 33.10 \\
                       & \citet{DBLP:conf/coling/VladikaSM24} & 68.24 & 66.84 & 67.53 & & - & - & - & & - & - & - \\ \hline
                       
                    \textsf{CER} (\textit{zero-shot})   & \textsf{CER} (DeBERTA v3-large) & 58.26 & 51.48 & 54.61 & & 93.10 & 90.48 & 91.77 & & 55.72 & 51.90 & 53.74
                        \\ \hdashline
                       
      \textsf{CER} (\textit{fine-tuned})              & \textsf{CER} (DeBERTA v3-large) & 64.55 & 70.14 & 67.22 & & \textbf{93.70} & \textbf{96.75} & \textbf{95.20} & & \textbf{61.44} & \textbf{60.72} & \textbf{61.14} \\
                       & \textsf{CER} (PubMedBERT) & 67.55 & \textbf{72.43} & \textbf{69.90} & & 89.76 & 90.02 & 89.88 & & 57.70 & 56.13 & 56.91 \\
                       & \textsf{CER} (BERT)       & \textbf{69.92} & 69.33 & 69.62 & & 81.28 & 83.89 & 82.56 & & 58.10 & 55.89 & 56.97 \\
      \hline
    \end{tabular}
  \end{adjustbox}
  \caption{\textbf{Comparison with state-of-the-art baselines}. This table presents the macro precision, recall, and F1 scores for different baseline models evaluated on the experimented datasets. Best results are highlighted in bold. '--' indicates instances where a technique could not be applied. }
  \label{tab:results_healthfc_bioasq}
\end{table*}

\subsection{Results} \label{sec: results}

\subsubsection{Comparison with state-of-the-art baselines}

\begin{table}[t]
    \centering
    \scalebox{0.99}{
    \begin{tabular}{llccc}
        \hline
        \textbf{Dataset} & \textbf{Scientific Evidence} & \textbf{P (\%)} & \textbf{R (\%)} & \textbf{F1 (\%)} \\
        \hline
        \multirow{2}{*}{\textbf{HealthFC}} & No & 40.10 & 41.24 & 40.61\\
                                           & Yes & \textbf{67.55} & \textbf{72.43} & \textbf{69.90} \\
        \hline
        \multirow{2}{*}{\textbf{BioASQ-7b}}   & No & 72.43 & 70.69 & 71.54 \\
                                           & Yes & \textbf{93.70} & \textbf{96.75} & \textbf{95.20} \\
        \hline
        \multirow{2}{*}{\textbf{SciFact}}  & No & 45.89 & 40.60 & 43.02 \\
                                           & Yes & \textbf{61.44} & \textbf{60.72} & \textbf{61.14} \\
        \hline
    \end{tabular}}
    \caption{\textbf{Impact of Scientific Evidence Retrieval.} This table shows the performance of \textsf{CER} with and without the scientific evidence retrieved from the PubMed knowledge base. P, R and F1 denote the macro precision, recall and F1 scores, respectively.}
    \label{tab: RAG_Evaluation}
\end{table}

In this work, we evaluated our \textsf{CER} framework in both zero-shot and fine-tuned configurations with state-of-the-art baselines. We used several pre-trained classification models at the foundation of \textsf{CER}, including general-domain models (BERT, DeBERTa-v3) and a domain-specific biomedical model (PubMedBERT). Results presented in Table \ref{tab:results_healthfc_bioasq} show that \textsf{CER} achieves consistent improvements across all datasets, with notable enhancements in F1 score and overall performance. Specifically, on the HealthFC dataset, fine-tuned \textsf{CER} achieves an F1 score of 69.90\%, outperforming both its zero-shot configuration (54.61\%) and the other models, emphasizing the benefits of fine-tuning on task-specific data. The results on BioASQ-7b further reinforce this trend: \textsf{CER}, fine-tuned, reaches a F1 score of 95.20\%, surpassing all the other baselines. Notably, \textsf{CER}'s zero-shot performance on BioASQ-7b (F1 = 91.77\%) remains competitive, indicating that it can leverage pre-existing knowledge from general-domain pretraining, which is crucial for zero-shot tasks where no additional labeled data is available. On the SciFact dataset, the fine-tuned \textsf{CER} model achieves an F1 score of 61.14\%, while its zero-shot counterpart obtains an F1 score of 53.74\%. Despite not being fine-tuned, the zero-shot version still outperforms all other baseline models. Interestingly, PubMedBERT, despite being pre-trained on biomedical data, performs slightly worse than \textsf{CER} on SciFact (F1 = 56.91\%) and BioASQ-7b (F1 = 89.88\%), suggesting that \textsf{CER} benefits from the combination of general-domain pretraining and fine-tuning, which may improve generalization across multiple biomedical domains. DeBERTa-v3’s strong performance on BioASQ-7b (F1 = 95.20\%) and SciFact (F1 = 61.14\%) is attributed to its advanced transformer architecture, which excels in handling more complex linguistic patterns, but fine-tuned \textsf{CER} consistently outperforms it on HealthFC and BioASQ-7b. Overall, the fine-tuned results for \textsf{CER} across HealthFC (F1 = 69.90\%), BioASQ-7b (F1 = 95.20\%), and SciFact (F1 = 61.14\%) underscore its ability to generalize well across a range of datasets. The competitive zero-shot performance across all datasets further highlights the strength of \textsf{CER} in effectively leveraging both general-domain and biomedical pretraining for downstream tasks. These findings emphasize the importance of fine-tuning for specialized tasks while also demonstrating the utility of combining pre-trained models for robust performance in zero-shot settings.

\subsubsection{Impact of Scientific Evidence Retrieval}
To quantify the contribution of the Scientific Evidence Retrieval module (Section \ref{sec: scientific_evidence}) to the overall \textsf{CER} framework, we conducted an ablation study comparing the full pipeline against a variant without retrieved context for justification generation and veracity classification. Table \ref{tab: RAG_Evaluation} demonstrates that removing scientific evidence retrieval leads to substantial performance degradation across all evaluation metrics, with macro-F1 scores decreasing by up to 29.3\%. \textit{The performance decline can be attributed to the lack of grounded scientific evidence and the consequent increased occurrence of hallucinated content in the LLM's reasoning process}. 

\footnotetext[6]{\href{https://www.perplexity.ai/}{Perplexity.ai website}}
\footnotetext[7]{\href{https://huggingface.co/mistralai/Mixtral-8x22B-Instruct-v0.1}{Mixtral-8x22B model on HuggingFace}}
\setcounter{footnote}{7}

\begin{table}[t]
    \centering
    \scalebox{0.99}{
    \begin{tabular}{llccc}
        \hline
        \textbf{Dataset} & \textbf{Retriever model} & \textbf{P (\%)} & \textbf{R (\%)} & \textbf{F1 (\%)} \\
        \hline
        \multirow{2}{*}{\textbf{HealthFC}} & Dense Retriever & 64.09 & 71.10 & 67.36 \\
                                           & Sparse Retriever & \textbf{67.55} & \textbf{72.43} & \textbf{69.90} \\
        \hline
        \multirow{2}{*}{\textbf{BioASQ-7b}}   & Dense Retriever & 92.77 & 96.47 & 94.61 \\
                                           & Sparse Retriever & \textbf{93.70} & \textbf{96.75} & \textbf{95.20} \\
        \hline
        \multirow{2}{*}{\textbf{SciFact}}  & Dense Retriever & \textbf{62.75} & \textbf{61.72} & \textbf{62.23} \\
                                           & Sparse Retriever & 61.44 & 60.72 & 61.14 \\
        \hline
    \end{tabular}}
    \caption{\textbf{Impact of Retriever Models.} This table shows the performance of Dense and Sparse Retriever models on three datasets: HealthFC, BioASQ-7b, and SciFact. P, R, and F1 denote the macro precision, recall, and F1 scores, respectively.}
    \label{tab: Retriever Evaluation}
\end{table}

\subsubsection{Evaluation of Dense vs Sparse Retrieval Approaches}
We evaluated both Dense and Sparse Retrieval approaches across the three experimental datasets. As shown in Table \ref{tab: Retriever Evaluation}, the difference in performance between the two methods was marginal across all datasets. Sparse Retrieval demonstrated a slight advantage on HealthFC (F1: 69.90\% vs. 67.36\%), while Dense Retrieval achieved marginally better performance on both BioASQ-7b (F1: 95.20\% vs. 95.08\%) and SciFact (F1: 62.23\% vs. 61.14\%). Interestingly, the consistent performance across both retrieval paradigms suggests that our \textsf{CER} framework's effectiveness is not heavily dependent on the choice of retrieval method. \textit{This robustness stems from the framework's justification mechanism, which appears to effectively leverage the retrieved scientific evidence regardless of the retrieval approach employed}. This finding has important implications for practical deployments, as it allows for flexible selection of retrieval methods based on computational constraints or specific domain requirements.

\subsubsection{Evaluation of Dense vs Sparse Retrieval Approaches}
We evaluated both Dense and Sparse Retrieval approaches across the three experimental datasets. As shown in Table \ref{tab: Retriever Evaluation}, the difference in performance between the two methods was marginal across all datasets. Sparse Retrieval demonstrated a slight advantage on HealthFC (F1: 69.90\% vs. 67.36\%), while Dense Retrieval achieved marginally better performance on both BioASQ-7b (F1: 95.20\% vs. 95.08\%) and SciFact (F1: 62.23\% vs. 61.14\%). Interestingly, the consistent performance across both retrieval paradigms suggests that our \textsf{CER} framework is relatively robust to the choice of retrieval method. \textit{This may be attributed to the justification mechanism, which is designed to synthesize relevant information from the retrieved scientific evidence, even when retrieval quality varies slightly.} This finding has important implications for practical deployments, as it allows for flexible selection of retrieval methods based on computational constraints or specific domain requirements.

\subsubsection{Impact of LLM Reasoning}
To evaluate the influence of prompt construction on the performance of LLM, we conducted experiments with various configurations to analyze how different reasoning factors affect prediction accuracy. 
This prompt --- designed using a Chain-of-Thought (CoT) \cite{DBLP:conf/nips/Wei0SBIXCLZ22} --- consists in three sections: (1) the \textit{<<SYS>>} section describes the LLM’s role and the tasks it needs to perform, (2) the \textit{<<Context>>} section provides sentences extracted from the knowledge base, equipping the LLM with the necessary information to classify the claim and justify its prediction, and (3) the \textit{<<Question>>} section presents the claim that the LLM is asked to assess. 

In Table \ref{tab: Justifications_Evaluation}, we present the results of an ablation study conducted using \textsf{CER} to evaluate the impact of different prompt components on LLM reasoning performance. Specifically, we investigate the performance when the prompt omits (1) the assignment of the \textit{Doctor} role, (2) the inclusion of scientific evidence, and (3) the requirement to provide a justification. Results show that the full \textsf{CER} prompt, incorporating the Doctor role assignment, scientific evidence, and justification requirement, achieves the highest performance in terms of Precision (P), Recall (R), and F1-score. In particular, on the HealthFC dataset, the \textsf{CER} prompt achieves an F1 score of 69.90\%, outperforming the ablated versions. Removing the Doctor role assignment results in a substantial drop to 33.26\% F1, highlighting the importance of framing the LLM as a medical expert. Similarly, omitting scientific evidence or the justification requirement leads to F1-scores of 40.13\% and 39.11\%, respectively. The BioASQ-7b dataset further reinforces this observation, but removing scientific evidence has a less drastic, though still noticeable, impact, reducing the F1-score to 91.63\%. On the SciFact dataset, the \textsf{CER} prompt achieves an F1-score of 61.14\%. Similar to the other datasets, ablating the Doctor role assignment, scientific evidence, or justification requirement leads to considerable performance degradation. These results support our hypothesis that \emph{combining scientific evidence with LLM reasoning, facilitated by explicit justifications and an appropriate role assignment, improves the accuracy of claim verification}.

The template used to generate justifications is shown below:

\begin{tcolorbox}[colback=yellow!10, colframe=yellow!80!red, title=\textit{LLM Reasoning} Prompt Template]
    \textbf{<<SYS>>}: \textit{You are a helpful, respectful, and honest Doctor. Always answer as helpfully as possible using the context text provided. Try to give an explanation based on the scientific evidence as follows: \textcolor{blue}{<Sentences returned by the Scientific Evidence module>}. The claim is as follows: \textcolor{blue}{claim}. Elaborate the scientific evidence to generate new information.}
    
    \textit{Use only the knowledge in the results to answer.}
    \textit{Be formal and use the third person.}
    \textit{Provide a maximum 200-word response without directly mentioning the Context. Use it but don’t refer to it in your answer. Begin with: ``Label:`` followed by the justification label (``Yes``, ``No``, or ``NEI``).}
    
    \textbf{<<Context>>}: \textit{Scientific evidence retrieved.}
    
    \textbf{<<Question>>}: \textit{Claim typed by the user.}
\end{tcolorbox}

\begin{table}[t]
    \centering
    \scalebox{0.90}{
    \begin{tabular}{llccc}
        \hline
        \textbf{Dataset} & \textbf{Prompt Construction} & \textbf{P (\%)} & \textbf{R (\%)} & \textbf{F1 (\%)} \\
        \hline
        \multirow{4}{*}{\textbf{HealthFC}} & \textsf{CER} & \textbf{67.55} & \textbf{72.43} & \textbf{69.90} \\
                                  & w/o \textit{Doctor} role assignment & 37.10 & 30.15 & 33.26 \\
                                  & w/o scientific evidence & 41.81 & 38.59 & 40.13 \\
                                  & w/o justification  & 40.38 & 37.93 & 39.11 \\
        \hline
        \multirow{4}{*}{\textbf{BioASQ-7b}} & \textsf{CER} & \textbf{93.70} & \textbf{96.75} & \textbf{95.20} \\
                                  & w/o \textit{Doctor} role assignment & 92.12 & 51.01 & 65.69 \\
                                  & w/o scientific evidence & 92.80 & 90.48 & 91.63 \\
                                  & w/o justification & 80.13 & 62.14 & 69.99 \\
        \hline
        \multirow{4}{*}{\textbf{SciFact}} & \textsf{CER} & \textbf{61.44} & \textbf{60.72} & \textbf{61.14} \\
                                  & w/o \textit{Doctor} role assignment & 30.87  & 27.08 & 28.85 \\
                                  & w/o scientific evidence & 39.65 & 34.49 & 36.89 \\
                                  & w/o justification & 42.01 & 40.64 & 41.31 \\
        \hline
    \end{tabular}}
    \caption{\textbf{Impact of LLM Reasoning.} This table shows the performance of \textsf{CER} on the experimented datasets by varying the prompt with a combination of sentences and justifications. The results are macro-averaged across experiments.}
    \label{tab: Justifications_Evaluation}
\end{table}

\subsubsection{Cross-dataset evaluations}
This experiment evaluated the \textsf{CER} architecture's generalization across biomedical datasets in a cross-dataset training and testing scenario. For the purpose of proper cross-dataset performance evaluation, given that BioASQ-7b contains only two classes (``true`` or ``false``), it was decided to introduce HealthFC-2 and SciFact-2, which are the original datasets previously encountered but without the ``NEI`` (Not Enough Information) class. Training on the diverse BioASQ-7b dataset improved generalization to specialized datasets (HealthFC-2, SciFact-2), demonstrating the importance of larger and more diverse datasets for cross-domain robustness. In contrast, training on specialized datasets was less effective for generalizing to BioASQ-7b. These results suggest that while the \textsf{CER} architecture shows some generalization ability, fine-tuning or transfer learning may further enhance performance, particularly when transitioning from specialized to broader datasets.

\begin{figure}[t]
    \centering
    % Prima riga di sottoparti
    \begin{subfigure}[b]{0.32\linewidth}  
        \centering
        \includegraphics[width=\linewidth]{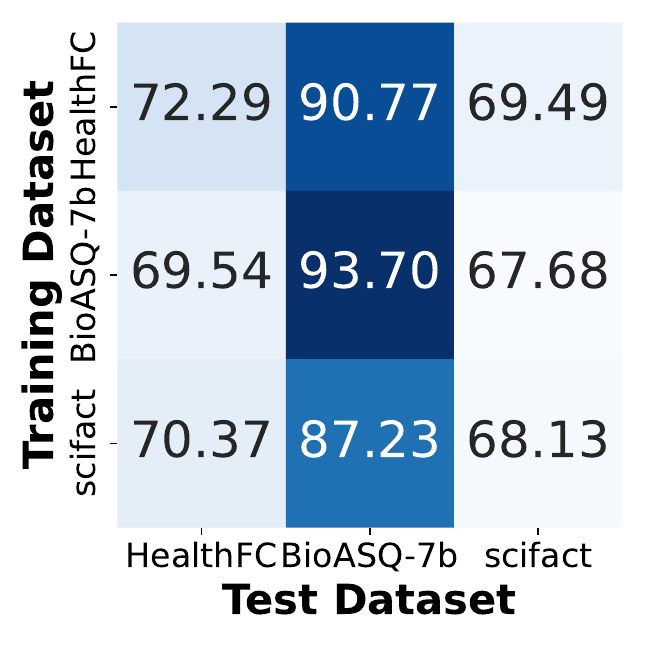}
        \caption{Precision}
        \label{fig:precision_matrix_1}
    \end{subfigure}
    \hfill
    \begin{subfigure}[b]{0.32\linewidth}  
        \centering
        \includegraphics[width=\linewidth]{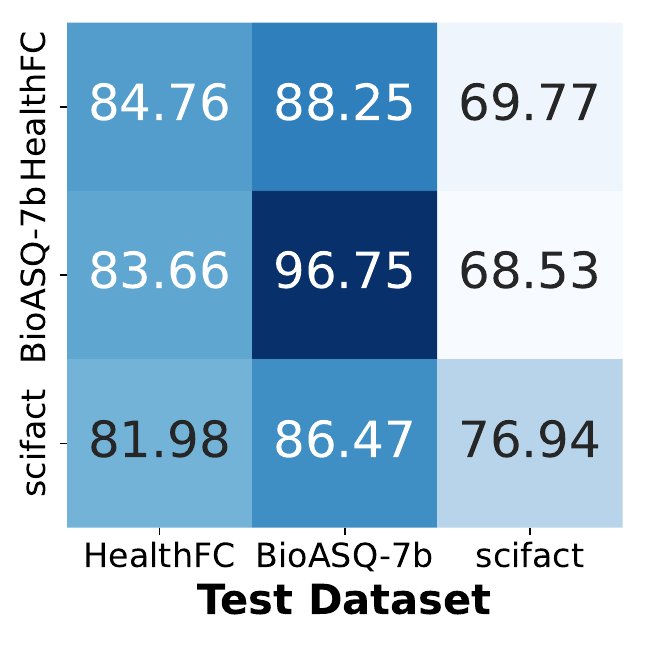}
        \caption{Recall}
        \label{fig:recall_matrix_1}
    \end{subfigure}
    \hfill
    \begin{subfigure}[b]{0.32\linewidth}  
        \centering
        \includegraphics[width=\linewidth]{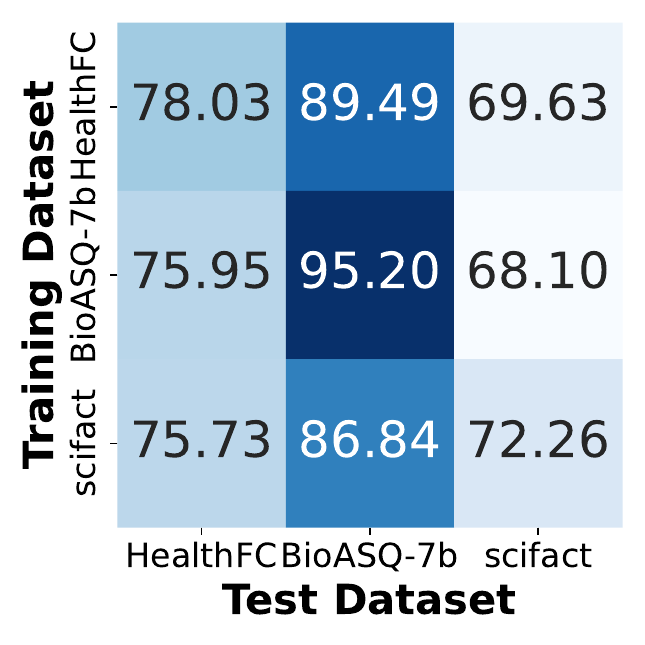}
        \caption{F1}
        \label{fig:f1_matrix_1}
    \end{subfigure}
    
    \caption{\textbf{Cross-dataset evaluation results.} Three matrix of macro precision, recall, and F1 scores for cross-dataset evaluations across the three experimental datasets.}
    \label{fig:compact_confusion_matrix}
    \Description{Cross-dataset evaluation compact layout.}
\end{figure}

\section{Related work} \label{sec: related_work}

Fact-checking has evolved through various approaches, including natural language analysis for claim verification \cite{Das_et_all}, evidence retrieval and matching from knowledge bases \cite{gautam2024factgeniuscombiningzeroshotprompting}, and machine learning models for classifying claims as true or false \cite{DBLP:conf/acl/AtanasovaSLA20, DBLP:conf/emnlp/AlhindiPM18, DBLP:journals/corr/abs-2406-05845}. The advent of Artificial Intelligence (AI) has introduced innovative methodologies to enhance these processes, making fact-checking more efficient and scalable. Comprehensive surveys, such as those by Guo et al. \cite{DBLP:journals/corr/abs-2108-11896} and Kotonya et al. \cite{DBLP:conf/emnlp/KotonyaT20}, analyze these techniques, identifying key research challenges and proposing frameworks for claim detection, evidence retrieval, and verification. These frameworks have catalyzed the development of online fact-checking tools like ClaimBuster \cite{DBLP:conf/kdd/HassanALT17}, Full Fact\footnote{\href{https://fullfact.org/}{Full Fact's website}}, The Factual (IsThisCredible?)\footnote{\href{https://credibilitycoalition.org/credcatalog/project/the-factual/}{The Factual's website}}, and Google’s Fact Check Explorer\footnote{\href{https://toolbox.google.com/factcheck}{Google's Fact Check Explorer website}}. Hartley et al. \cite{Hartley_2024} evaluated these tools, comparing their accuracy with independent fact-checking organizations and emphasizing their significance in combating misinformation. Initially developed to verify political claims and online content, fact-checking has become essential for maintaining information accuracy, particularly given its influence on public opinion and decision-making. Research into identifying linguistic markers of fake news, such as the work by Rashkin et al. \cite{DBLP:conf/emnlp/RashkinCJVC17}, has expanded to include automated fact-checking in social media contexts. Bhatnagar et al. \cite{DBLP:conf/coling/BhatnagarKC22} demonstrated the efficacy of both ``out-of-the-box`` and fine-tuned summarization models for these tasks. The importance of fact-checking has become particularly pronounced in the healthcare domain, where misinformation about treatments, diagnoses, and disease prevention can pose significant public health risks. During the COVID-19 pandemic, Mohr et al. \cite{DBLP:conf/lrec/MohrWK22} highlighted the detrimental effects of such misinformation. To address these challenges, researchers have adapted political fact-checking methodologies to healthcare, creating new datasets and techniques. For instance, Barve et al. \cite{DBLP:journals/tjs/BarveS23} developed an architecture employing a Content Similarity Vector to match URLs with reliable sources. Similarly, \citet{DBLP:conf/aaai/Gong0W0024, DBLP:conf/coling/KimC20} proposed a word-level heterogeneous graph to better capture semantic relationships between claims and evidence. Additionally, datasets like BioASQ-7b \cite{DBLP:journals/corr/abs-2006-09174}, HealthFC \cite{DBLP:conf/coling/VladikaSM24}, and SciFact \cite{wadden2020emnlp} have been essential for testing automated fact-checking systems in health domain, emphasizing the importance of using scientific evidence to make such systems more accurate as pointed out by Vladika et al. \cite{DBLP:journals/corr/abs-2404-08359}. Fine-tuned Transformer models have shown superior performance over large language models (LLMs) in tasks such as claim detection, while LLMs excel in generative tasks like question decomposition for evidence retrieval as described by Zhang et al. \cite{DBLP:conf/coling/ZhangG24}. However, LLMs face challenges such as hallucination, which can degrade their performance. To address this, Bèchard et al. \cite{DBLP:journals/corr/abs-2404-08189} recommended employing techniques like retrieval-augmented generation (RAG) to integrate external knowledge and mitigate these issues. Building on this foundation, our study presents an innovative framework that addresses the limitations of current LLM-based approaches in medical claim verification. By integrating structured knowledge bases with LLMs, our method grounds generated explanations in verified information, reducing issues like hallucination and enhancing reliability. Simultaneously, we leverage fine-tuned transformer models optimized for claim verification tasks to improve predictive accuracy. This hybrid approach effectively balances interpretability and precision, providing transparent, evidence-based insights crucial for safeguarding public health.

\section{Conclusion \& Future Work} \label{sec: conclusion}

In this work, we presented a novel approach to biomedical fact-checking, named \textsf{CER} (Combining Evidence and Reasoning). Our system leverages the strengths of both scientific evidence retrieval and LLM reasoning to achieve robust and accurate claim verification. We evaluated \textsf{CER} on three publicly available datasets and demonstrated its effectiveness through consistent improvements over prior state-of-the-art methods. Furthermore, we conducted an ablation study to analyze the impact of different design choices and shed light on the interplay between evidence integration and LLM reasoning. Cross-dataset evaluations confirm \textsf{CER}'s strong generalization ability, crucial for applications across diverse biomedical domains. Future work could extend \textsf{CER}'s evidence retrieval to additional biomedical databases for richer context and improved accuracy. Efforts should also focus on enhancing domain generalization through adaptive training or creating diverse datasets encompassing broader biomedical topics and claim types.

\paragraph{Limitations} The architecture relies on pre-trained language models for the generation of justifications. These models have known limitations, including a tendency to generate hallucinated content (incorrect but plausible-sounding information) \cite{DBLP:journals/corr/abs-2310-01469}, particularly when dealing with zero-shot tasks. While our pipeline mitigates this issue with the supervised \textit{Veracity Prediction} step, generating justifications with models that are free from the hallucinations problem could potentially improve prediction accuracy.

%\paragraph{Ethical Considerations} This paper implements a fact-checking system for medical claims. Although promising results, they are not yet ready for everyday use due to incorrect predictions caused by hallucinations, lack of information, and difficulty in interpreting models, which could lead to harmful consequences and further spread misinformation.

\section*{Acknowledgements}
This work was conducted with the financial support of (1) the PNRR MUR project PE0000013-FAIR and (2) the Italian ministry of economic development, via the ICARUS (Intelligent Contract Automation for Rethinking User Services) project (CUP: B69J23000270005).

\begin{comment}
\begin{figure}[ht]
    \centering
    \begin{subfigure}{0.6\textwidth} 
        \centering
        \includegraphics[width=1\textwidth]{img/Appendix/2.pdf}
        \caption{Example from the HealthFC dataset with generated justification and label predicted using fine-tuning.}
        \label{fig:NEI_example}
    \end{subfigure}
    \hfill
    \begin{subfigure}{0.6\textwidth} 
        \centering
        \includegraphics[width=1\textwidth]{img/Appendix/3.pdf}
        \caption{Example from the BioASQ dataset with generated justification and label predicted using fine-tuning.}
        \label{fig:Yes_example}
    \end{subfigure}
    \caption{Examples from the HealthFC and BioASQ datasets with generated justifications and labels predicted using fine-tuning.}
    \label{fig:CER_Examples}
    \Description{CER examples}
\end{figure}
\end{comment}

\balance
\appendix

\begin{table*}[t!]
    \centering
    \scalebox{0.90}{
    \begin{tabular}{p{0.2\textwidth}p{0.86\textwidth}}
        \hline
        \textbf{LLM} & \textbf{Justification} \\
        \hline
        \textbf{Mixtral} \cite{jiang2024mixtralexperts} & Recent studies have established a link between bites from the Lone Star Tick (Amblyomma americanum) and the development of red meat allergy. This allergy is mediated by the production of IgE antibodies against the carbohydrate galactose-$\alpha$-1,3-galactose ($\alpha$-gal), which is present in the tick's saliva. The tick bite triggers the production of these antibodies, leading to a hypersensitivity reaction upon subsequent consumption of red meat. This phenomenon has been observed globally, with the Lone Star Tick identified as a common vector. Therefore, it is accurate to state that a bite from the Lone Star Tick can cause the victim to become allergic to red meat. \\
        \hline
        \textbf{M42-health/Llama3-Med42-8B} \cite{med42v2} & Lone Star Tick - Amblyomma Americanum (pdf) (cdc.gov) Lone Star Tick (Amblyomma americanum) - Vector of Choice for Rickettsial Diseases (vets4pets.com) Lone Star Tick (Amblyomma americanum) - Animal Bites and M.D. Facts (paws.org) CDC - Tick Size Comparison (cdc.gov) Lone Star Tick - Amblyomma americanum \\
        \hline
        \textbf{BioGPT} \cite{10.1093/bib/bbac409} & \textit{Repeat the prompt without generating a justification} \\
        \hline
        \textbf{Healix-1.1B-V1-Chat-dDPO} \footnotemark[11] & \textit{Continue repeating 'Answer: Yes' without providing any context or justification, even if the prompt does not include information about the label or a predicted answer.}
        \\
        \hline
        \textbf{Writer/palmyra-med-70b-32k} \cite{palmyra} & The context mentions that the rat pancreatitis-associated protein (pap) is a 16 kDa secretory protein. This indicates that pap is a protein that is secreted from cells, as it is found in the pancreatic juice after induction of pancreatic inflammation. The context also states that pap is synthesized as a preprotein with a predicted molecular weight of 16,630, further supporting the idea that it is a secretory protein. Therefore, based on the information provided, it can be concluded that pap is a secreted protein.\\
        \hline
    \end{tabular}}
    \caption{Example of different justifications generated for the same claim and evidences using different LLMs.}
    \label{tab:Justification-2}
\end{table*}

\section{Appendix}

\subsection{Fine-tuning parameters} \label{sec:appendix-A}

To determine the optimal parameter values, we designed a structured experimental plan based on best practices from the literature on similar prediction tasks and tested which parameter combinations yielded the best results \cite{j2024finetuningllmenterprise}. We tested learning rate values ranging from $2\cdot e^{-5}$ to $4\cdot e^{-5}$, finding that $2\cdot e^{-5}$ offered the best balance between convergence and performance. Batch size for training and evaluation were set to 8 to optimize memory usage and model performance (per device). Several epoch values were tested, with 5 epochs proving sufficient for convergence without overfitting. A weight decay of 0.20 helped prevent overfitting while maintaining model generalization. Gradient accumulation steps have been set to 2, allowing for larger effective batch sizes without exceeding memory constraints. The metric used to select the best model was the macro F1-score, and the save strategy was based on epochs.

\subsection{Healthcare LLM} \label{sec:appendix-B}

We also generated justifications using other LLMs trained in the medical field to determine if their specialization could improve the generated justifications. The LLMs selected for this analysis were 'bioGPT' \cite{10.1093/bib/bbac409}, 'm42-health/Llama3-Med42-8B' \cite{med42v2}, 'Healix-1.1B-V1-Chat-dDPO'\footnote{\href{https://huggingface.co/health360/Healix-1.1B-V1-Chat-dDPO}{Healix-1.1B model on HuggingFace}}, and 'writer/palmyra-med-70b-32k' \cite{palmyra}.

These LLMs are smaller than Mixtral, which is used in the \textsf{CER} architecture. They are less powerful, and most failed to accurately grasp the context and information retrieved from PubMed, as shown in the justifications in Table \ref{tab:Justification-2}. The only LLM capable of generating justifications comparable to Mixtral was ``writer/palmyra-med-70b-32k`` \cite{palmyra}. Consequently, we proceeded to generate results following the \textsf{CER} architecture using the justifications generated by this model, as shown in Table \ref{tab:palmyra}.

\begin{table}[t]
    \centering
    \scalebox{0.8}{
    \begin{tabular}{lcccccccccc}
        \hline
        \textbf{Fact-Checker} & \multicolumn{3}{c}{\textbf{HealthFC}} & & \multicolumn{3}{c}{\textbf{BioASQ-7b}} \\
        \cline{2-4} \cline{6-8}
        & \textbf{P (\%)} & \textbf{R (\%)} & \textbf{F1 (\%)} & & \textbf{P (\%)} & \textbf{R (\%)} & \textbf{F1 (\%)} \\
        \hline
        \textsf{CER} (DeBERTa-v3) & 46.59 & 48.66 & 47.26 & & \textbf{75.25} & \textbf{76.26} & \textbf{75.64}  \\
        \textsf{CER} (PubMedBERT) & 55.27 & 56.66 & \textbf{55.32} & & 68.41 & 68.66 & 68.53\\
        \textsf{CER} (BERT) & \textbf{56.11} & \textbf{58.00} & 53.31 & & 53.33 & 40.00 & 45.71 \\
        \hline
    \end{tabular}}
    \caption{Results obtained using the Palmyra LLM, which was implemented specifically for justification creation.}
    \label{tab:palmyra}
\end{table}

\subsection{LLM Reasoning output examples} \label{sec:appendix-C}

Table \ref{tab:HealthFC} and \ref{tab:BioASQ-7b} show additional examples of LLM Reasoning outputs on claims extracted from HealthFC and BioASQ-7b, respectively.

\begin{table}[t]
    \centering
    \scalebox{0.7}{
    \begin{tabular}{lp{0.55\textwidth}}
        \toprule
        \textbf{Dataset} & HealthFC \\
        \midrule
        \textbf{Claim} & Does pine wood improve sleep? \\
        \midrule
        \textbf{Justification} & \textit{The context provided does not contain specific information regarding the direct impact of pine wood on sleep improvement. While it mentions the long-lasting odor of certain pine species that has been reported to relieve stress in humans, it does not establish a direct link between pine wood and improved sleep. Therefore, based on the given context, it is not possible to definitively state that pine wood improves sleep. More specific research would be needed to confirm this claim.} \\
        \bottomrule
    \end{tabular}}
    \caption{Evaluation of the claim from the HealthFC dataset.}
    \label{tab:HealthFC}
\end{table}

\begin{table}[t]
    \centering
    \scalebox{0.7}{
    \begin{tabular}{lp{0.55\textwidth}}
        \toprule
        \textbf{Dataset} & BioASQ-7b \\
        \midrule
        \textbf{Claim} & Can capsules or tablets with lactase improve the symptoms of lactose intolerance and make dairy products more tolerable? \\
        \midrule
        \textbf{Justification} & \textit{The administration of lactase supplements, such as capsules or tablets, has been shown to improve lactose digestion in individuals with impaired or absent intestinal lactase activity. This is supported by a randomized, placebo-controlled, double-blind, 5-arm crossover study, which demonstrated that different lactase preparations and freeze-dried yogurt culture can affect gastrointestinal lactose digestion and reduce symptoms of lactose intolerance. Furthermore, microbial-derived beta-galactosidase enzyme preparations have been shown to improve in vivo lactose digestion and tolerance through enhanced gastrointestinal digestion of lactose. Therefore, it can be concluded that capsules or tablets with lactase can improve the symptoms of lactose intolerance and make dairy products more tolerable.} \\
        \bottomrule
    \end{tabular}}
    \caption{Evaluation of the claim from the BioASQ-7b dataset.}
    \label{tab:BioASQ-7b}
\end{table}

\newpage
%% The next two lines define the bibliography style to be used, and the bibliography file.
\bibliographystyle{ACM-Reference-Format}
\balance{}
\bibliography{CER}
\end{document}